\documentclass[journal,twoside,web]{ieeecolor}
\usepackage{tmi}
\usepackage{cite}
\usepackage{amsmath,amssymb,amsfonts}
\usepackage{algorithmic}
\usepackage{graphicx}
\usepackage{textcomp}
\def\BibTeX{{\rm B\kern-.05em{\sc i\kern-.025em b}\kern-.08em
    T\kern-.1667em\lower.7ex\hbox{E}\kern-.125emX}}
\usepackage{multirow}
\usepackage{makecell}
\usepackage{hyperref}
\usepackage[switch]{lineno}

\newcommand{\etal}{\textit{et al}.}
\newcommand{\ie}{\textit{i}.\textit{e}.}
\newcommand{\eg}{\textit{e}.\textit{g}.}

\makeatletter
\def\fnum@figure{\textcolor{subsectioncolor}{\sf Fig.~\thefigure}}
\def\fnum@table{\textcolor{subsectioncolor}{\sf TABLE~\thetable}}
\makeatother

\begin{document}
\title{Contour Transformer Network for One-shot Segmentation of Anatomical Structures}

\author{Yuhang Lu, Kang Zheng, Weijian Li, Yirui Wang, Adam P. Harrison, Chihung Lin, Song Wang, Jing Xiao, Le Lu, Chang-Fu Kuo, and Shun Miao
\thanks{Y. Lu and S. Wang are with the Department of Computer Science and Engineering, University of South Carolina,
Columbia, SC 29208, USA. (e-mail: yuhang@email.sc.edu).}
\thanks{K. Zheng, Y. Wang, A. P. Harrison, L. Lu and S. Miao are with PAII Inc., Bethesda, MD 20817, USA. (e-mail: miaoshun638@paii-labs.com).}
\thanks{W. Li is with the Department of Computer Science, University of Rochester, Rochester, NY 14627, USA.}
\thanks{C. Lin and C. Kuo are with Chang Gung Memorial Hospital, Linkou, Taiwan, ROC. (e-mail: zandis@gmail.com)}
\thanks{J. Xiao is with Ping An Technology, Shenzhen, CO 518029,  China.}
\thanks{This work was mainly done when Y. Lu and W. Li were interns at PAII Inc.}
}

\maketitle

\begin{abstract}
Accurate segmentation of anatomical structures is vital for medical image analysis. The state-of-the-art accuracy is typically achieved by supervised learning methods, where gathering the requisite expert-labeled image annotations in a scalable manner remains a main obstacle.
Therefore, annotation-efficient methods that permit to produce accurate anatomical structure segmentation are highly desirable. 
In this work, we present \textit{Contour Transformer Network} (CTN), a one-shot anatomy segmentation method with a naturally built-in human-in-the-loop mechanism.
We formulate anatomy segmentation as a contour evolution process and model the evolution behavior by graph convolutional networks (GCNs).
Training the CTN model requires only one labeled image exemplar and leverages additional unlabeled data through newly introduced loss functions that measure the global shape and appearance consistency of contours.
On segmentation tasks of four different anatomies, we demonstrate that our one-shot learning method significantly outperforms non-learning-based methods and performs competitively to the state-of-the-art fully supervised deep learning methods. With minimal human-in-the-loop editing feedback, the segmentation performance can be further improved to surpass the fully supervised methods.
\end{abstract}


\section{Introduction}
\IEEEPARstart{S}{egmentation} of anatomical structures serves as a core element in a wide spectrum of medical image analysis applications. Recent advances in deep learning research have significantly boosted the accuracy of medical image segmentation. However, without abundant pixel-level labels, the state-of-the-art segmentation methods~\cite{ronneberger2015u,dolz2018hyperdense,RothLLHFSS18,wang2020deep,chen2018encoder,harrison2017progressive} cannot achieve their optimal performance~\cite{tajbakhsh2020embracing}.
Annotating segmentation masks for medical images is extremely time-consuming and requires specialized expertise on human anatomy and its variations. 
As a result, prompt solutions are demanded to train an accurate segmentation model with limited labeled data.

\begin{figure}[t]
	\begin{center}
		\includegraphics[width=\linewidth]{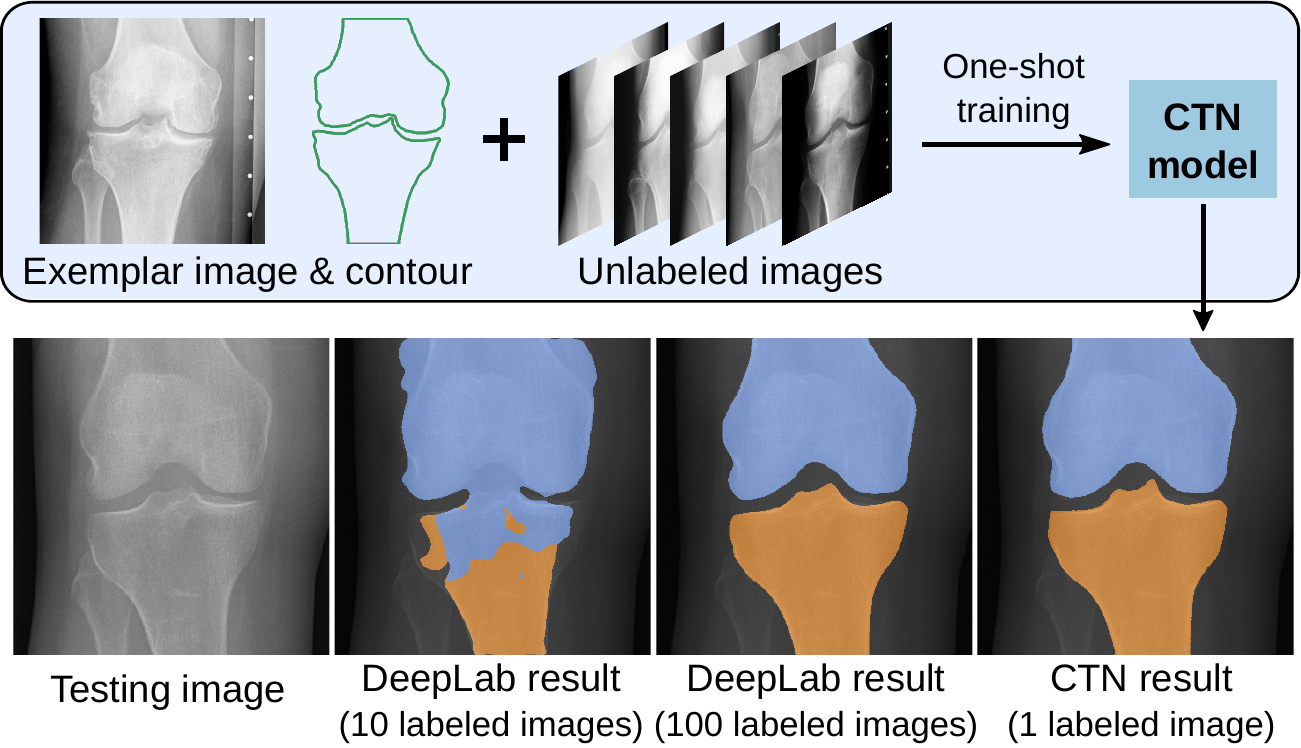}
	\end{center}
	\caption{\textbf{An overview of CTN}. CTN could learn to segment the anatomical structure accurately from only one exemplar and a set of unlabeled images. In contrast, fully supervised methods such as DeepLab~\cite{chen2018encoder} will fail when training with insufficient labeled images.}
	\label{fig:intro}
\end{figure}

One-/few-shot image segmentation methods have been studied in recent years aiming to reduce the dependency on large labeled data. Knowledge transfer is widely adopted for one-/few-shot segmentation of natural images~\cite{shaban2017one,michaelis2018one,dong2018few,zhang2019canet}. 
These methods leverage on external labeled datasets (e.g., PASCAL VOC~\cite{Everingham10} and MS-COCO~\cite{lin2014microsoft}) to learn general knowledge of segmentation and is able to transfer the knowledge to object categories given a small labeled support set.
Although the object category to be segmented is not seen during training, a large labeled dataset of diversified objects is still required. In the medical image domain, especially plain X-ray, such a labeled dataset is still not available yet.
More important, there is still a significant accuracy gap between existing one-/few-shot methods and fully supervised ones.

In this work, we propose an annotation-efficient anatomical structure segmentation method, termed \textit{Contour Transformer Network} (CTN). Our work is inspired by the human annotator's capability of learning segmentation of anatomical structure from one or very few exemplars. This is achieved by understanding the shape and appearance traits of the target object from the exemplars and actively looking for objects with similar traits in new images. To mimic this behavior, we propose a semi-supervised learning approach that exploits the shape and appearance similarities of the target object between labeled and unlabeled images to train a segmentation model. As a result, CTN is able to learn segmentation from one labeled exemplar and a set of unlabeled images without dependency on external labeled datasets (Fig.~\ref{fig:intro}).


Owing to the inherent regularized nature of anatomical structures, the same anatomy in different (X-ray) images may share common features or properties, such as the anatomical structure's \textit{shape}, \textit{appearance} and \textit{gradients} along the structural object boundary. Although different images are not directly comparable, we can compare their common features only and use the exemplar segmentation to guide other unlabeled images partially, thus making CTN trainable in a one-shot setting. 
Specifically, we formulate the segmentation problem as learning a contour evolution behavior modeled by a cascaded graph convolutional network (GCN). Three differentiable contour-based loss functions namely \textit{contour perceptual loss}, \textit{contour bending loss} and \textit{edge loss} are proposed to describe the common features of appearance, shape and edge response, respectively. For each unlabeled image, CTN takes the exemplar contour as an initialization, then gradually evolves it under the guidance from the three losses. We evaluated CTN on four X-ray image segmentation tasks and demonstrated that it significantly outperforms previous one-shot segmentation methods and performs competitively when compared to fully supervised methods.



An efficient \textit{human-in-the-loop} mechanism is a compelling feature for one-/few-shot segmentation in applications demanding extreme precision, e.g., measuring the joint space in X-rays. However, existing one-/few-shot methods often lack such a mechanism, leaving an accuracy gap that renders them unfit for many accuracy-critical applications. In contrast, CTN has a native human-in-the-loop mechanism that allows its performance to be improved by learning from annotation-efficient corrections. Namely, we format manual corrections as partial contours where users need to only redraw incorrectly segmented parts and leave correct parts untouched. These partial contour annotations can be naturally incorporated back into the training via an additional Chamfer loss~\cite{barrow1977parametric}. 
We demonstrate that with minimum human-in-the-loop feedback, CTN can outperform fully supervised methods on all four X-ray datasets evaluated. 

In summary, our contributions are four-fold: 
1) We propose CTN, a one-shot anatomical structure segmentation method that can be trained using one exemplar and a set of unlabeled images, without depending on external labeled data.
2) We propose two new differentiable loss functions \textit{contour perceptual loss} and \textit{contour bending loss}, plus the existing \textit{edge loss}, to enable GCNs to integrate anatomical priors of appearance, shape and gradient, respectively.
3) We design a human-in-the-loop mechanism to allow CTN to utilize additional manual labels with low annotation cost.
4) We demonstrate on four datasets that CTN achieves the state-of-the-art one-shot segmentation results, i.e., it performs competitively when compared to fully supervised alternatives and outperforms them with minimal human-in-the-loop feedback.

A preliminary version of this work has been published in
a conference proceeding~\cite{lu2020learning}. In this paper, we made the following extensions:
1) we add evaluations on a new dataset of hip X-ray images and provide more result analysis and discussion; 2) We conduct new experiments to further analyze the behavior of the proposed method, including evaluations with more unlabeled images, different loss weights and different exemplar images, analysis on failure and corner cases; 3) We add more comprehensive discussion on the relationship/comparison between our work and related work, more detailed technical description of the proposed method and in-depth discussion of the limitations and our future work.


\subsection{Related Work}

\subsubsection{Non-learning-based segmentation}
Classic segmentation methods include solutions based on directly optimizing a pre-defined energy function. 
Well known examples include level-set~\cite{cremers2007review}, active contour model (ACM)~\cite{kass1988snakes}, graph-cut~\cite{boykov2001fast}, random walker~\cite{grady2006random} and their variants~\cite{caselles1997geodesic,chan2001active,marquez2013morphological}.
Although classic methods have limited performance and are no longer state-of-the-art, their essential concepts and philosophy remain insightful. 
We adopt the contour evolution scheme from ACM by representing segmentation using contours. 
Instead of optimizing the gradient-based energy function on individual images to obtain a segmentation, we optimize compound losses concerning shape, appearance, and gradient on the whole training set to learn a contour evolution policy.

Atlas and multi-atlas methods can also perform segmentation task given only one or a few examples~\cite{baillard2001segmentation,ciofolo2009atlas,iglesias2015multi}. 
However, the required image registration is a challenging task by itself~\cite{oliveira2014medical}, and inter-subject image appearance variance can lead to inaccurate registration and segmentation~\cite{zhao2019data}.

\subsubsection{Supervised segmentation}
State-of-the-art supervised learning based segmentation methods are predominantly using deep learning, specifically fully convolutional network (FCN)~\cite{long2015fully} and its variants~\cite{ronneberger2015u,dolz2018hyperdense,wang2020deep,chen2018encoder}. 
These methods follow a per-pixel classification framework, where each pixel is classified individually by the deep neural network. 
Lacking constraints from a global structure, deep learning segmentation methods typically require a large number of labeled images to be trained effectively. 
When training data quantities are insufficient, the performance tends to degrade significantly, as shown in Fig.~\ref{fig:intro}.

Incorporating anatomical priors into neural network training has been proven useful in recent studies. \cite{oktay2017anatomically} employs a shape regularization autoencoder in a segmentation network to constrain the prediction to follow a learned shape distribution. 
\cite{lee2019tetris} takes a shape template as an additional input channel and deforms it to match the underlying structure through a spatial transformer network. 
While these methods exploit shape prior to improve segmentation robustness, they still require a large number of labeled images. 
In contrast, CTN exploits the shape and appearance commonality between labeled and unlabeled images to achieve one-shot segmentation.

\begin{figure*}[htbp]
	\begin{center}
		\includegraphics[width=\linewidth]{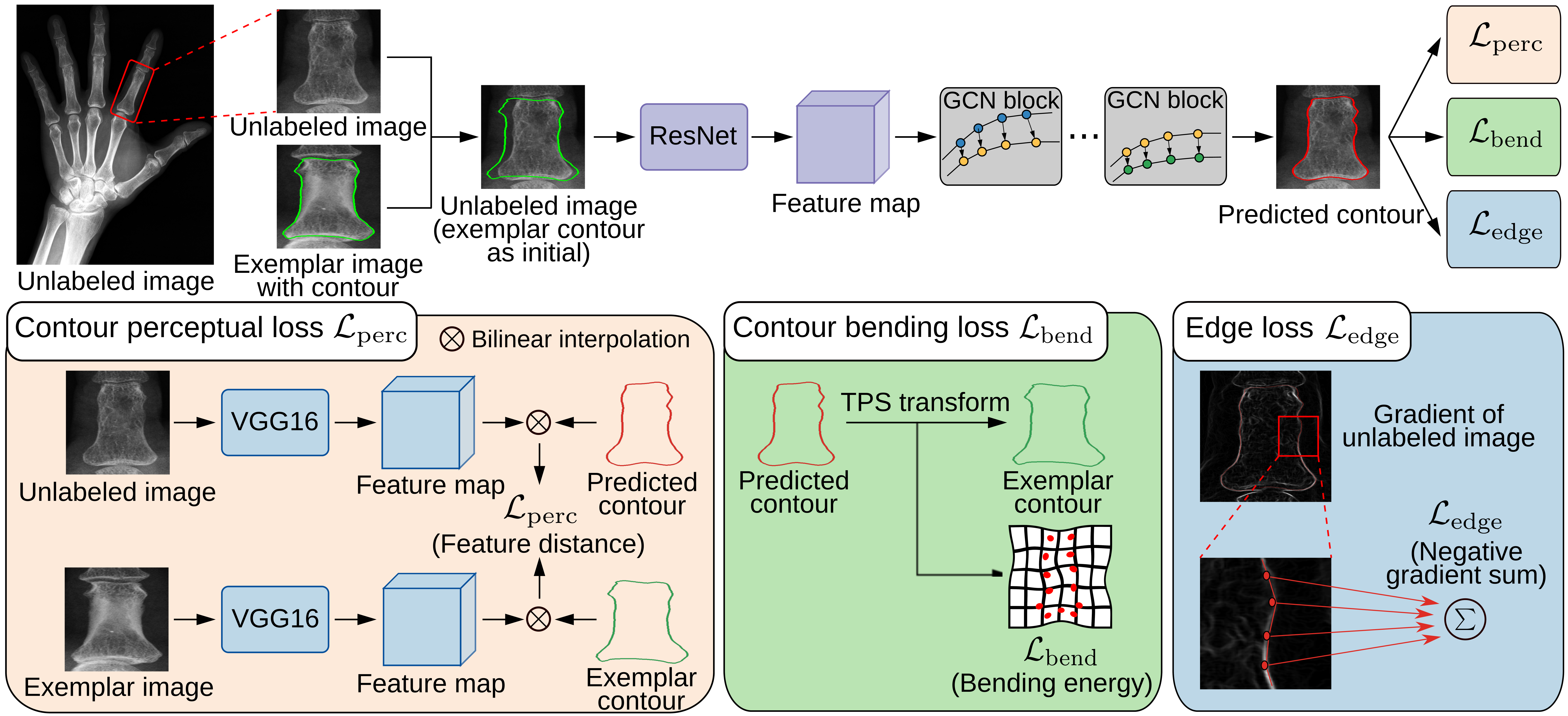}
	\end{center}
	\caption{\textbf{Contour Transformer Network.} CTN is trained to fit a contour to the object boundary by learning from one exemplar. In training, it takes a labeled exemplar and a set of unlabeled images as input. After going through a CNN encoder and five GCN contour evolution blocks, it outputs the predicted contour. We train the network using three one-shot losses (\ie, contour perceptual loss, contour bending loss and edge loss), aiming to let the predicted contour have similar contour features with the exemplar.}
	\label{fig:framework}
\end{figure*}

Learning-based ACMs have also been studied to segment a variety of objects including heart~\cite{rupprecht2016deep, chen2019learning}, blood vessel~\cite{gur2019unsupervised} and building~\cite{gur2019unsupervised, marcos2018learning}. These methods incorporate deep learning and ACM by learning the ACM energy terms \cite{marcos2018learning} or evolution directions ~\cite{rupprecht2016deep}, constructing a ACM-inspired network architecture \cite{gur2019unsupervised} and loss \cite{chen2019learning}. While CTN also learns the contour evolution policy, it differs from ~\cite{rupprecht2016deep} by introducing novel losses to support one-shot learning and employs GCN to allow effective information exchange along the contour.

\subsubsection{One-/few-shot segmentation} 
One-/few-shot segmentation methods aim to segment objects of a new category learned from a small support set of labeled examples~\cite{wang2019generalizing}. 
Existing works on natural images~\cite{shaban2017one,michaelis2018one,dong2018few,zhang2019canet} mostly leverage the pre-training on a large and comprehensive annotated dataset like MS-COCO~\cite{lin2014microsoft}.
This condition renders the above approach inapplicable to the medical image domain, where such equivalent large labeled datasets simply do not exist, especially when considering a collection of specialized anatomical structure and imaging modality to be addressed. 

Data augmentation is another common approach to solve this problem in medical imaging. Various generative models have been used in recent works to generate synthetic training data, such as variational autoencoders~\cite{dalca2018anatomical}, generative adversarial networks~\cite{tang2019ct,wang2020lt} and transformation networks~\cite{zhao2019data}.
A comprehensive survey of using imperfect datasets in medical image segmentation can be found in~\cite{tajbakhsh2020embracing}. 
Zhao \etal~\cite{zhao2019data} propose an approach to model both spatial and appearance transformations between images in the entire dataset, and synthesize images with the learned transformations for training segmentation models.
Our work is partially inspired by the same motivation, but instead of learning a data augmentation model, we exploit the inherent regularized nature of anatomical structures, using one exemplar to guide the segmentation.

\begin{figure*}[htbp]
	\begin{center}
		\includegraphics[width=0.8\linewidth]{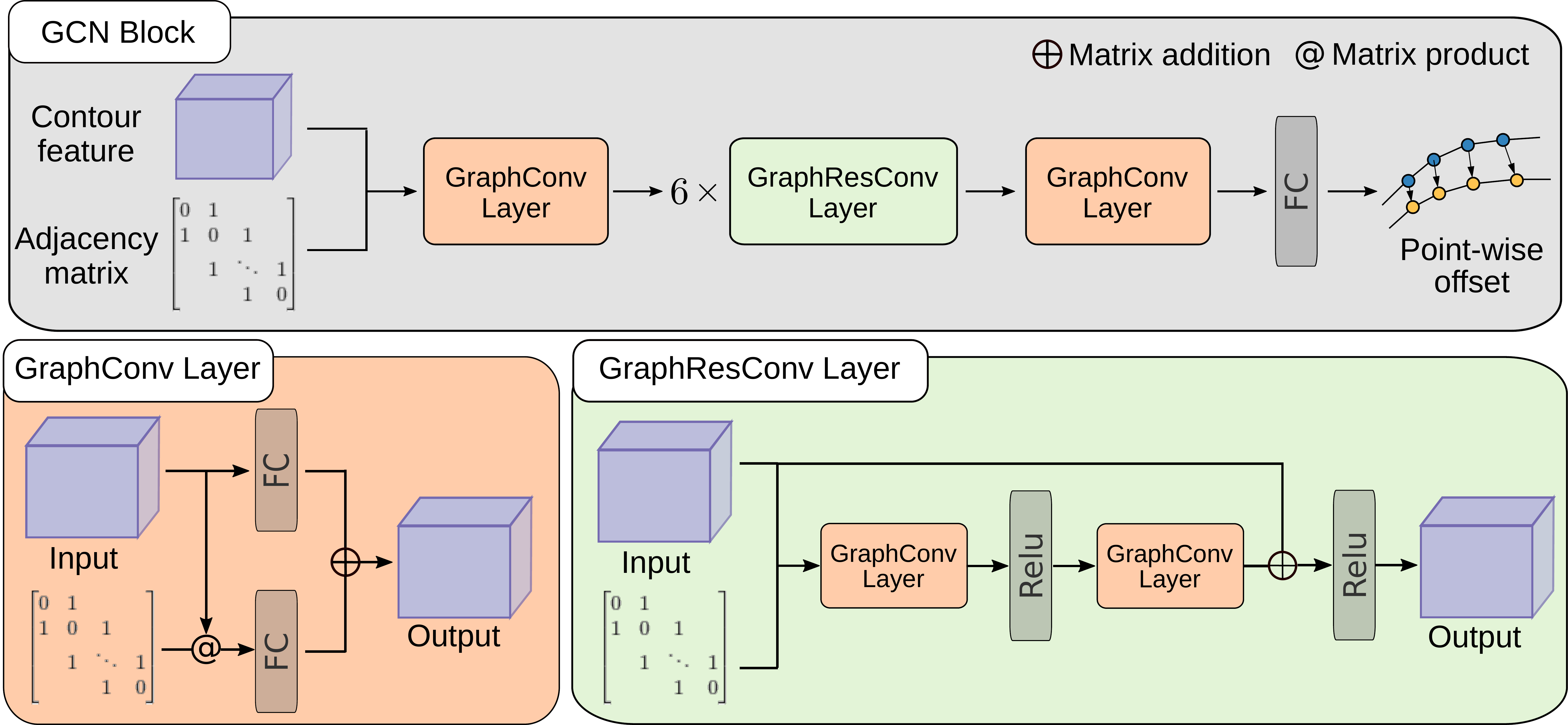}
	\end{center}
	\caption{\textbf{Network Architecture of GCN blocks.} CTN uses five cascaded GCN blocks to model the contour evolution behavior. They take image features along the contour and an adjacency matrix that represents vertex connections as input, and predict point-wise offsets to update the contour. Their architectures are identical, but weights are not shared.}
	\label{fig:gcn_blcok}
\end{figure*}

\section{Method}

\subsection{Overview} \label{sec:overview}
The problem of anatomical structure segmentation can be decomposed into two steps: ROI (Region of Interest) detection; and ROI segmentation.
ROI detection can be achieved via landmark detection and has been well-studied in past literature~\cite{wu2018look,chen2019cephalometric,wang2020deep,li2020structured}, so we focus on achieving very high segmentation accuracy by taking the detected ROI (with noise and errors) as input images. 

The training pipeline of CTN is illustrated in Fig.~\ref{fig:framework}. Our task is to learn an segmentation model of an anatomical structure from a set of unlabeled images $\{I\}$ and an exemplar image $I_E$ with its segmentation $C_E$ of the target structure. We model each segmentation as a contour, represented by a fixed number of evenly spaced vertices, $C=\{\mathbf{p}_1, \mathbf{p}_2, \dots, \mathbf{p}_N\}$. For each unlabeled image $I$, its contour $C$ is initialized by placing the exemplar contour $C_E$ at the center of the image. CTN models the contour evolution policy that displaces the initial contour $C$ to the boundary of the target structure in $I$. It can be written as: 
\begin{equation}
    F_{\boldsymbol{\theta}}(I_E, C_E, I, C) = \Delta C
\end{equation}
where $F_{\boldsymbol{\theta}}$ denotes the CTN with weights $\boldsymbol{\theta}$. It takes the exemplar and the target image as input, and outputs estimated offsets of contour vertices.

Due to the lack of labels on $I$, fully supervised losses cannot be used to train CTN. Here, we exploit the advantage of modeling segmentation as contour, \ie, it provides natural representations of the segmentation's boundary and shape. In particular, instead of comparing model predictions with ground truth as in a fully supervised setting, we compare $C$ with the exemplar contour $C_E$, by measuring the dissimilarities between their shapes and the local image patterns along with them. This is motivated by the insight that the correct segmentation in the target image should be similar to the exemplar contour in its overall shape, as well as local image appearance patterns of corresponding vertices. As a side benefit, the predicted contours of CTN are naturally corresponded to the exemplar contour.

We propose two new losses to measure the shape and appearance dissimilarities: namely contour perceptual loss, denoted as $L_{perc}$, and contour bending loss, denoted as $L_{bend}$. In addition, we employ the classic gradient-based loss, denoted as $L_{edge}$, to further drive the contour to edges. Details of these losses will be described in Section~\ref{sec:loss}. CTN is trained by minimizing weighted combination of the three losses:
\begin{equation} \label{eq:one_shot_loss}
    \min_{\boldsymbol{\theta}} \sum_{\{I\}} \lambda_1 \cdot  L_{perc} + \lambda_2 \cdot L_{bend} + \lambda_3 \cdot L_{edge}
\end{equation}
where $\lambda_1, \lambda_2, \lambda_3$ are weighting factors of the three losses. An illustration of the training process of CTN is shown in Fig.~\ref{fig:framework}.

These losses imitate the human's behavior in learning contouring from one exemplar, \ie, drawing new contours by referring to the exemplar to compare shapes and local appearances. Another key insight is that although these losses can be used in an ACM setting (where the contour vertices are directly optimized to minimize the energy), training CTN on aggregating over the entire unlabeled dataset is robust, stable and can inhibit the boundary leaking issue on individual cases often encountered by ACM.

\subsection{Network architecture}
\label{sec:arch}

Following~\cite{ling2019fast}, we use a CNN-GCN architecture to model contour evolution. As shown in Fig.~\ref{fig:framework}, CTN consists of two parts: an image encoding CNN block and subsequent cascaded contour evolution GCN blocks. ResNet-50~\cite{he2016deep} is employed as the backbone of the image encoding block. It takes the target image as input and outputs a feature map encoding local image appearances, denoted as:
\begin{equation}
    f = F_{cnn}(I).
\end{equation}

All contour evolution blocks have the same multi-layer GCN structure, although weights are not shared. The GCN takes the contour graph with vertex features as input, denoted as $G=(C, E, Q)$, where $C$ denotes the contour vertices, $E$ denotes the connectivity, and $Q$ denotes the vertex features. Each vertex in the contour is connected to four neighboring vertices, two on each side. The vertex features are extracted from the feature map $f$ at vertex locations via bilinear interpolation, which can be written as:
\begin{equation}
Q = \{ f(\mathbf{p}) \}_{\mathbf{p} \in C}
\label{eqn:gfeature}
\end{equation}
where $f(\mathbf{p})$ denotes the result of bilinear interpolation of $f$ at location $\mathbf{p}$.

 Five GCN blocks are cascaded to evolve the contour. The $k$-th block takes the graph $G_k=(C_k, E, Q_k)$ as input, and outputs offsets of the contour vertices:
\begin{equation}
    C_{k+1} = C_k + F^k_{gcn}(C_k, E, Q_k).
\end{equation}
The contour is initialized using the exemplar contour, $C_0 = C_E$, and the output of the last contour evolution block is the final output.



The architecture of GCN blocks is shown in Fig~\ref{fig:gcn_blcok}. Each GCN block consists of 2 graph convolutional (GraphConv) layers~\cite{kipf2016semi}, 6 graph residual convolutional (GraphResConv) layers~\cite{pixel2mesh} and 1 fully connected (FC) layer. The first GraphConv layer and all GraphResConv layers have 256 channels. The last GraphConv layer has 32 channels. The FC layer has 2 channels outputting the offsets on x and y axis, respectively. 



\subsection{One-shot training losses} \label{sec:loss}

\subsubsection{Contour perceptual loss} \label{sec:percep_loss}
We propose a contour perceptual loss to measure
the dissimilarity between the visual patterns of the exemplar contour $C_E$ on the exemplar image $I_E$ and the predicted contour $C$ on the target image $I$. Partially enlightened by the perceptual loss~\cite{johnson2016perceptual} developed for image super-resolution, which measures image perceptual similarities in the feature space of VGG-Net~\cite{vgg}, we measure contour perceptual similarities in the graph feature space. In particular, graph features are extracted from the VGG-16 feature maps of the two images along the two contours (similar to Eq.~\ref{eqn:gfeature}), and their L1 distance is calculated as the contour perceptual loss:
\begin{equation} \label{eq:perceptual_loss}
    L_{perc} = \sum_{i=1,\dots,N} \| P(\mathbf{p}_i) - P_E(\mathbf{p}'_i) \|_1
\end{equation}
where $\mathbf{p}_i \in C$, $\mathbf{p}'_i \in C_E$, and $P$ and $P_E$ denote the VGG-16 features of $I$ and $I_E$, respectively.
Following~\cite{johnson2016perceptual}, we use features at the layers of \texttt{relu1\_2}, \texttt{relu2\_2}, \texttt{relu3\_3}, and \texttt{relu4\_3} in VGG-16.
We first downscale the contour vertices using the downsample factor of each feature map, and then sample contour features from feature maps using bilinear interpolation, and last the contour features of four layers are concatenated for comparison.
The VGG-16 weights are pretrained on ImageNet~\cite{imagenet}.

Instead of using L2 distance found in the original perceptual loss formulation~\cite{johnson2016perceptual}, we employ L1 distance since it empirically performed better in our experiments. Because of the inevitable appearance variations across images, we hypothesize that the similarity representation between pairs of local image patterns is often limited according to certain aspects, \eg, specific texture, context, or shape features. Given that different channels of VGG-16 features capture different characteristics of local image patterns, a distance metric learning with modeling flexibility to select which salient features to match is more appropriate. The sparsity-inducing nature of L1 distance definition provides additional ``selection'' mechanism over L2, which may explain the improved performance observed. 

Using the contour perceptual loss to measure appearance similarity between contours has a few advantages: 1) Since VGG-16 network features can capture the image pattern of a neighboring area with spatial contexts (\ie, network receptive field), the contour perceptual loss enjoys a relatively large capturing range (\ie, the convex region around the minimum), making the CTN training optimization easier; 2) The backbone VGG-16 model is trained on ImageNet ~\cite{imagenet} for classification tasks, so that its learned features are more sensitive to underlying structure and less sensitive to noises and illumination variations, which improves the robustness of CTN training.

\subsubsection{Contour bending loss} \label{sec:shape_loss}

If we operate under the assumption that an exemplar contour is broadly informative to other data samples, then it should be beneficial to use the exemplar shape to ground any predictions on such other samples. To this end, we propose a contour bending loss to measure the shape dissimilarity between contours. The loss is calculated as the bending energy of the TPS warping~\cite{bookstein1989principal} that maps $C_E$ to $C$. It is worth noting that TPS warping achieves the minimum bending energy among all warpings that map $C_E$ to $C$.
Since bending energy measures the magnitude of the 2nd order derivatives of the warping, the contour bending loss penalizes more on local and acute shape changes, which are often associated with mis-segmentation.

Given a predicted contour $C$, the TPS bending energy can be calculated as follows:
\begin{align}
    \mathbf{K} &= \left ( \left \| \mathbf{p}^{\prime}_i - \mathbf{p}^{\prime}_j \right \|_2 ^ 2 \cdot log\left \| \mathbf{p}^{\prime}_i - \mathbf{p}^{\prime}_j \right \|_2 \right ) \\
    \mathbf{P} &= (\mathbf{1}, \mathbf{x}^{\prime}, \mathbf{y}^{\prime}) \\
    \mathbf{L} &= \begin{bmatrix} \mathbf{K} & \mathbf{P}\\ \mathbf{P}^T & \mathbf{0}\end{bmatrix}
\end{align}
where $\mathbf{p}_i=(x_i, y_i)$, $\mathbf{p}'_i=(x'_i, y'_i)$ are points of $C$ and $C_E$, respectively. $\mathbf{x}^{\prime}=\{x^{\prime}_1, x^{\prime}_2, \ldots, x^{\prime}_N\}^T$,  $\mathbf{y}^{\prime}=\{y^{\prime}_1, y^{\prime}_2, \ldots, y^{\prime}_N\}^T$. $\mathbf{K}$, $\mathbf{P}$,  $\mathbf{L}$ are matrices of size $N \times N$, $N \times 3$ and $(N+3) \times (N+3)$, respectively. Finally, the TPS bending energy is written as
\begin{equation} 
\mathcal{L}_{bend} = \max \left [ \frac{1}{8\pi}(\mathbf{x}^T \mathbf{H} \mathbf{x} + \mathbf{y}^T \mathbf{H} \mathbf{y}) ,0 \right]
\end{equation}
where $\mathbf{x}=\{x_1, x_2, \ldots, x_N\}^T$, $\mathbf{y}=\{y_1, y_2, \ldots, y_N\}^T$, and $\mathbf{H}$ is the $N \times N$ upper left submatrix of $\mathbf{L}^{-1}$~\cite{wang2017correspondence}.

\subsubsection{Edge loss}

Although the contour perceptual and bending losses can achieve robust segmentation, they are inherently insensitive to (very) small segmentation fluctuations, such as deviations from the correct boundary by a few pixels. Therefore, in order to obtain desirably high segmentation accuracies to adequately facilitate the downstream workflows like rheumatoid arthritis quantification~\cite{huo2015automatic}, we also employ an edge loss measuring the image gradient magnitude along the contour, which attracts the contour toward edges in the image. The edge loss is written as:
\begin{equation} 
\mathcal{L}_{edge} = - \frac{1}{N} \sum_{\mathbf{p} \in C} {\left \| \nabla I(\mathbf{p}) \right \|_2}
\end{equation}
where $\nabla$ is the gradient operator.

\begin{figure}[t]
	\begin{center}
		\includegraphics[width=\linewidth]{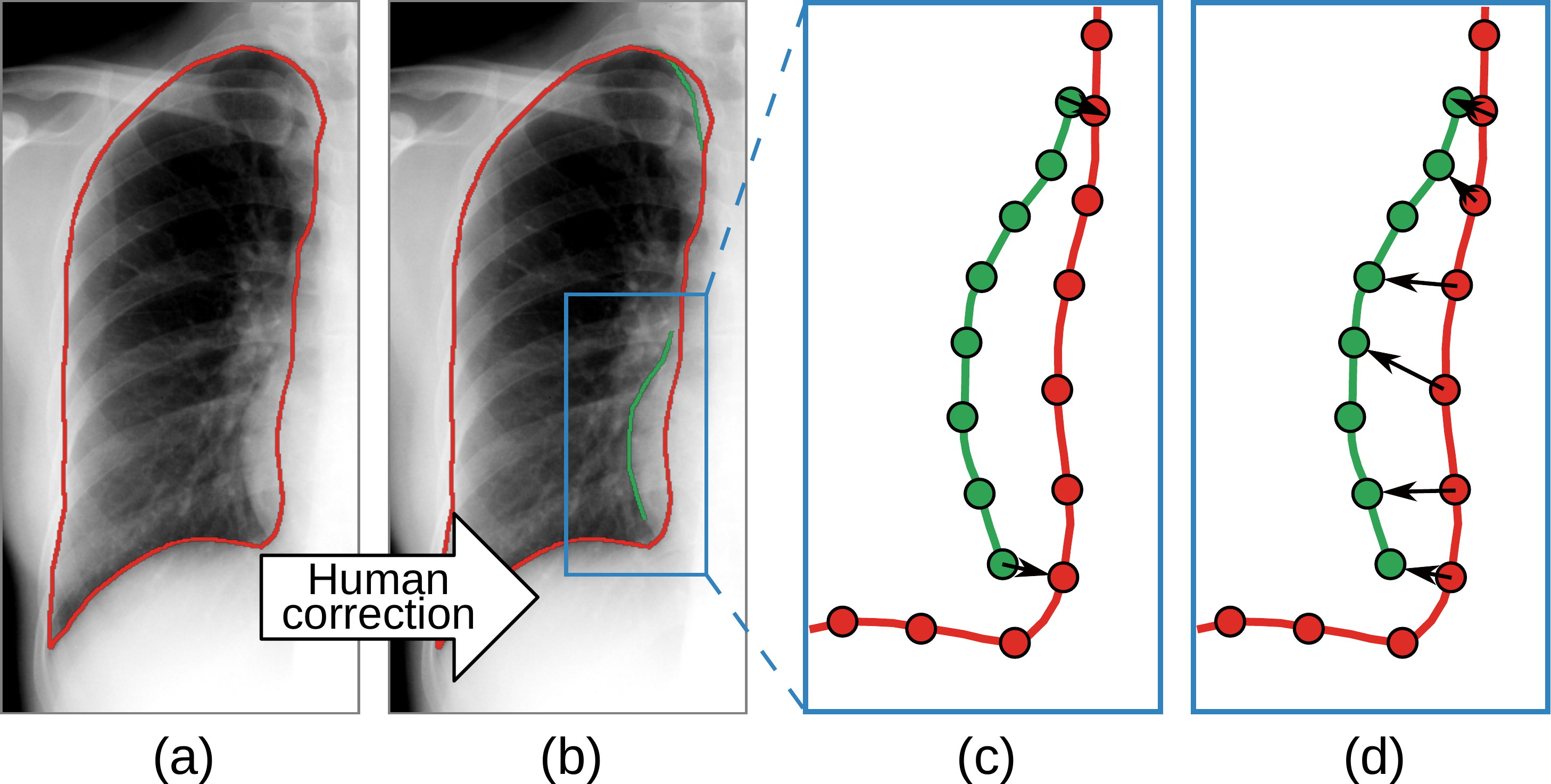}
	\end{center}
	\caption{\textbf{Human-in-the-loop}. Given a {\bf red} predicted contour (a), the annotator corrects its wrong parts with {\bf green} curves (b). For each corrected contour segment, we find two points in the predicted contour, closest to its start and end (c), then each predicted point between the two points are assigned to the closest corrected point (d). This prevents the point correspondence to be scattered.}
	\label{fig:human_in_the_loop}
\end{figure}

\begin{figure*}
	\begin{center}
		\includegraphics[width=\linewidth]{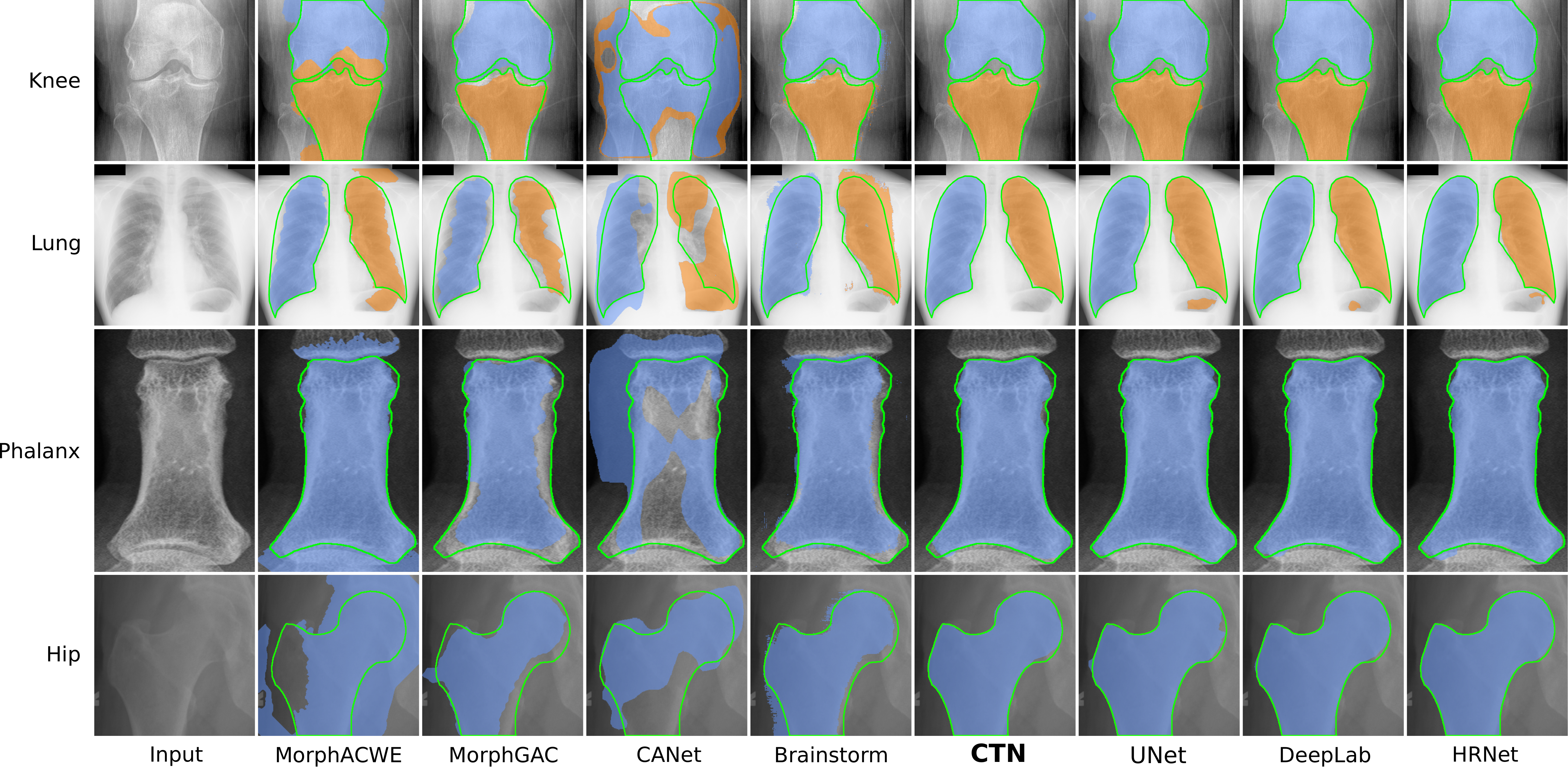}
	\end{center}
	\caption{\textbf{Segmentation results of four example images.} The boundaries of ground truth segmentations (the green lines) are drawn for comparison.}
	\label{fig:seg_compare}
\end{figure*}

\subsection{Human-in-the-loop} \label{sec:human}
Learning from one exemplar is based on the assumption that the anatomical structure has similar boundary features in all images. 
It works in most cases, but outliers are inevitable. 
To achieve even higher accuracy in testing, sometimes we need to consider more possibilities in training.
To this end, the proposed CTN offers a natural way to incorporate additional labeled images with a human-in-the-loop mechanism.

Assuming a CTN model is trained with one exemplar, we want to finetune it with more segmentation annotations. We first run this model on a set of unlabeled images and select a number of images with wrong predictions as new samples. Instead of drawing the whole contour from scratch on these new images, the annotator only needs to draw some partial contours, in order to correct the wrong prediction (as shown in Fig.~\ref{fig:human_in_the_loop}(b). The point-wise training of CTN makes it possible to learn from these partial corrections. This way, we reduce the labor cost to the minimum.

A \textit{partial contour matching loss} is proposed to utilize the partial ground truth contours during the CTN training. Denote $\hat{\mathbf{C}}$ as a set of partial contours in image $I$, each element of which is an individual contour segment. For each contour segment $\hat{C}_i \in \hat{\mathbf{C}}$, we build the point correspondence between $\hat{C}_i$ and $C$. For each $\hat{C}_i$, we find two points in the predicted contour $C$ that are closest to the start and end points of $\hat{C}_i$, then each predicted point between the two points are assigned to the closest corrected point. Denote the corresponding predicted contour segment by $C_i$ ($C_i \in C$). We define the distance between $C$ and $\hat{C}_i$ as the Chamfer distance from $C_i$ to $\hat{C}_i$:
\begin{equation}
D(\hat{C}_i, C) = \sum_{\mathbf{p} \in C_i}\min_{\mathbf{\hat{p}} \in \hat{C}_i}\left \| \mathbf{p}- \mathbf{\hat{p}} \right \|_2
\end{equation}
and the partial matching loss of $C$ is defined as:
\begin{equation}
L_{pcm} = \frac{1}{N}\sum_{\hat{C}_i \in \hat{\mathbf{C}}} D(\hat{C}_i, C).
\end{equation}

In the human-in-the-loop scenario, we combine all losses to train the CTN, and rewrite the Eq.~\ref{eq:one_shot_loss} as:
\begin{equation} \label{eq:total_loss}
    \min_{\boldsymbol{\theta}} \sum_{\{I\}} \lambda_1 \cdot  L_{perc} + \lambda_2 \cdot L_{bend} + \lambda_3 \cdot L_{edge} + \lambda_4 \cdot L_{pcm}
\end{equation}
which allows CTN to be trained with fully labeled, partially labeled and unlabeled images simultaneously and seamlessly. Whenever new labeled image are available, we can use Eq.~(\ref{eq:total_loss}) to finetune the existing CTN model.

\section{Experiments}

\setlength{\tabcolsep}{0.5mm}
\begin{table*}
\caption{Performances of CTN and eight existing methods on four datasets.} 
\begin{center}
\begin{tabular}{c|c|c c|c c|c c|c c|c c}
\hline
\multicolumn{2}{c|}{\multirow{2}{*}{Methods}}                    & \multicolumn{2}{c|}{Knee}                                & \multicolumn{2}{c|}{Lung}                                & \multicolumn{2}{c|}{Phalanx}                                 & \multicolumn{2}{c|}{Hip}                                 & \multicolumn{2}{c}{\textbf{Mean}} \\ \cline{3-12} 
\multicolumn{2}{c|}{}                                           & IoU (\%)                          & HD (px)                        & IoU (\%)                          & HD (px)                        & IoU (\%)           & HD (px)                                  & IoU (\%)           & HD (px)                               & IoU(\%)           & HD(px)          \\ \hline
\hline
\multirow{2}{*}{\makecell{Non-learning-\\based}}   & MorphACWE~\cite{marquez2013morphological}                       & 65.89$\pm$6.07                      & 54.07$\pm$3.77                     & 76.09$\pm$6.39                      & 55.35$\pm$17.82                     & 74.33$\pm$6.49       & 69.13$\pm$10.66      & 48.05$\pm$4.70       & 94.11$\pm$9.90       & 66.09       & 68.17      \\ 
                              & MorphGAC~\cite{marquez2013morphological}                        & 87.42$\pm$1.87                      & 15.78$\pm$3.02                     & 70.79$\pm$4.16                      & 45.67$\pm$6.92                     & 82.15$\pm$5.15       & 24.73$\pm$7.21      & 83.42$\pm$4.43       & 32.20$\pm$10.44 & 80.95       & 29.60            \\ \hline
\hline
\multirow{3}{*}{\makecell{One-shot}}     & CANet~\cite{zhang2019canet}                           & 29.22$\pm$3.63                       & 175.86$\pm$9.74                    & 56.90$\pm$7.09                      & 73.46$\pm$12.03                     & 60.90$\pm$7.02       & 67.13$\pm$7.09      & 48.89$\pm$16.26       & 88.39$\pm$23.35      & 48.98       & 101.21      \\ 
                              & Brainstorm~\cite{zhao2019data}                      & 90.17$\pm$1.72                      & 29.07$\pm$5.32                     & 77.13$\pm$4.71                      &43.28$\pm$8.38                      & 80.05$\pm$5.17       & 30.30$\pm$6.90        & 82.48$\pm$3.18       & 44.17$\pm$9.29        & 82.46       & 36.71        \\ 
                              & \textbf{CTN (Ours)}                      & \textbf{97.32$\pm$0.67}             & \textbf{6.01$\pm$1.42}             & \textbf{94.75$\pm$1.97}             & \textbf{12.16$\pm$5.87}            & \textbf{96.96$\pm$1.29} & \textbf{8.19$\pm$4.49}       & \textbf{97.29$\pm$0.72} & \textbf{8.27$\pm$3.06} & \textbf{96.58} & \textbf{8.66}       \\ \hline
\hline
\multirow{3}{*}{\makecell{Fully \\supervised}}   
                              & UNet~\cite{ronneberger2015u}                            & 96.60$\pm$1.61                      & 7.14$\pm$4.24                      & 95.38$\pm$1.87                      & 12.48$\pm$6.40                     & 96.76$\pm$1.76       & 10.10$\pm$6.84      & 96.51$\pm$4.22       & 13.28$\pm$14.55      & 96.31       & 10.75      \\ 
                              & DeepLab~\cite{chen2018encoder}                     & 97.18$\pm$0.67                      & 5.41$\pm$2.27                      & 96.18$\pm$1.40                      & 10.81$\pm$6.26                     & 97.63$\pm$0.93       & 6.52$\pm$3.32       & 97.64$\pm$0.72       & 6.24$\pm$2.63      & 97.16       & 7.25      \\ 
                              & HRNet~\cite{wang2020deep}                           & 96.99$\pm$0.65                      & 5.18$\pm$2.52                      & 95.99$\pm$1.39                      & 10.44$\pm$6.03                     & 97.47$\pm$1.31       & 7.03$\pm$4.43       & 97.66$\pm$2.38       & 7.57$\pm$6.71      & 97.03       & 7.56      \\ \hline
\end{tabular}
\end{center}
\label{table:compare}
\end{table*}

\subsection{Datasets and experimental settings}
\subsubsection{Datasets}
\label{sec:dataset}
We evaluate our method on four X-ray image datasets focusing on different anatomical structures of knee, lung, phalanx and hip, respectively.

\begin{itemize}
\item \textbf{Knee:} We randomly selected 212 knee X-ray images from the Osteoarthritis Initiative (OAI) database~\footnote{\url{https://nda.nih.gov/oai/}}. Each knee image is cropped from the original scan with automatic knee joint detection, and resized to $360 \times 360$ pixels. The dataset is randomly split into 100 training and 112 testing images.

\item \textbf{Lung:} We use the public JSRT dataset~\cite{JSRTdatabase} with 247 posterior-anterior chest radiographs, where lung segmentation labels originate from the SCR dataset~\cite{SCRdataset}~\footnote{\url{https://www.isi.uu.nl/Research/Databases/SCR/}}. 
Left lung and right lung ROIs are extracted from the image and resized to $512 \times 256$ pixels. Following~\cite{SCRdataset}, the 124 images with odd indices are used for training, and the 123 images with even indices for testing. 

\item \textbf{Phalanx:} We collected an in-house dataset of hand X-ray images from patients with rheumatoid arthritis. 202 ROIs of proximal phalanx are extracted from images automatically based on hand joint detection~\cite{huo2015automatic} and resized to $512 \times 256$ pixels. We randomly split the dataset into 100 training and 102 testing images.

\item \textbf{Hip:} We randomly selected 300 pelvic X-ray images from the OAI database, 100 for training and 200 for testing. Each hip image is cropped from the original scan with automatic landmark detection, and resized to $360 \times 360$ pixels.

\end{itemize}

On the knee, phalanx and hip datasets, we manually annotated the target objects, namely tibia, femur, phalanx and hip bones, under the guidance of a senior rheumatologist. The image lists and annotations of the knee and hip datasets are publicly available~\footnote{\url{https://github.com/rudylyh/CTN_data}}. For the knee and lung segmentation tasks, where there are multiple objects to be segmented, we train separate CTNs to segment the objects.

For every dataset, we selected the most representative image in the training set as the exemplar image based on the distance to other images. Specifically, for every image in the training set, we calculate its distance to all other images in the ImageNet-trained VGG feature space, which represents the semantic similarity between the two images. The image with minimum average distance to other images is selected as the exemplar.

\subsubsection{Evaluation metrics}
For each segmentation result, we evaluate segmentation accuracy by IoU and for the corresponding object contour by the Hausdorff distance (HD). For methods that do not explicitly output object contours, we extract the external contour of the largest region of each class from the segmentation mask. On the knee dataset, we report the average HD of femur and tibia segmentation.


\subsubsection{Implementation details}
The hyper-parameter settings are $N = 1000$, $\lambda_1 = 1$, $\lambda_2 = 0.25$, $\lambda_3 = 0.1$, $\lambda_4 = 1$. The network is trained using the Adam optimizer with a learning rate of $1\times10^{-4}$, a weight decay of $1\times10^{-4}$ and a batch size of 12 for 500 epochs. We use the same hyper-parameter setting in both one-shot training and human-in-the-loop finetuning.

\subsection{Comparison with existing methods} \label{sec:cmp_perform}

We compare CTN against seven representative methods from three categories: non-learning-based, one-shot, and fully supervised segmentation methods. The quantitative results are reported in Table~\ref{table:compare} and visualizations of segmentation results are shown in Fig.~\ref{fig:seg_compare}.
\subsubsection{Comparison with non-learning-based methods}
We first compare with two non-learning-based methods: MorphACWE~\cite{caselles1997geodesic, marquez2013morphological} and MorphGAC~\cite{chan2001active, marquez2013morphological}~\footnote{\url{https://github.com/pmneila/morphsnakes}}. Both of them are based on ACM, which evolves an initial contour to the object by minimizing an energy function. We use the exemplar contour of our method as their initial contours.

\begin{figure}
    \begin{center}
        \includegraphics[width=0.95\linewidth]{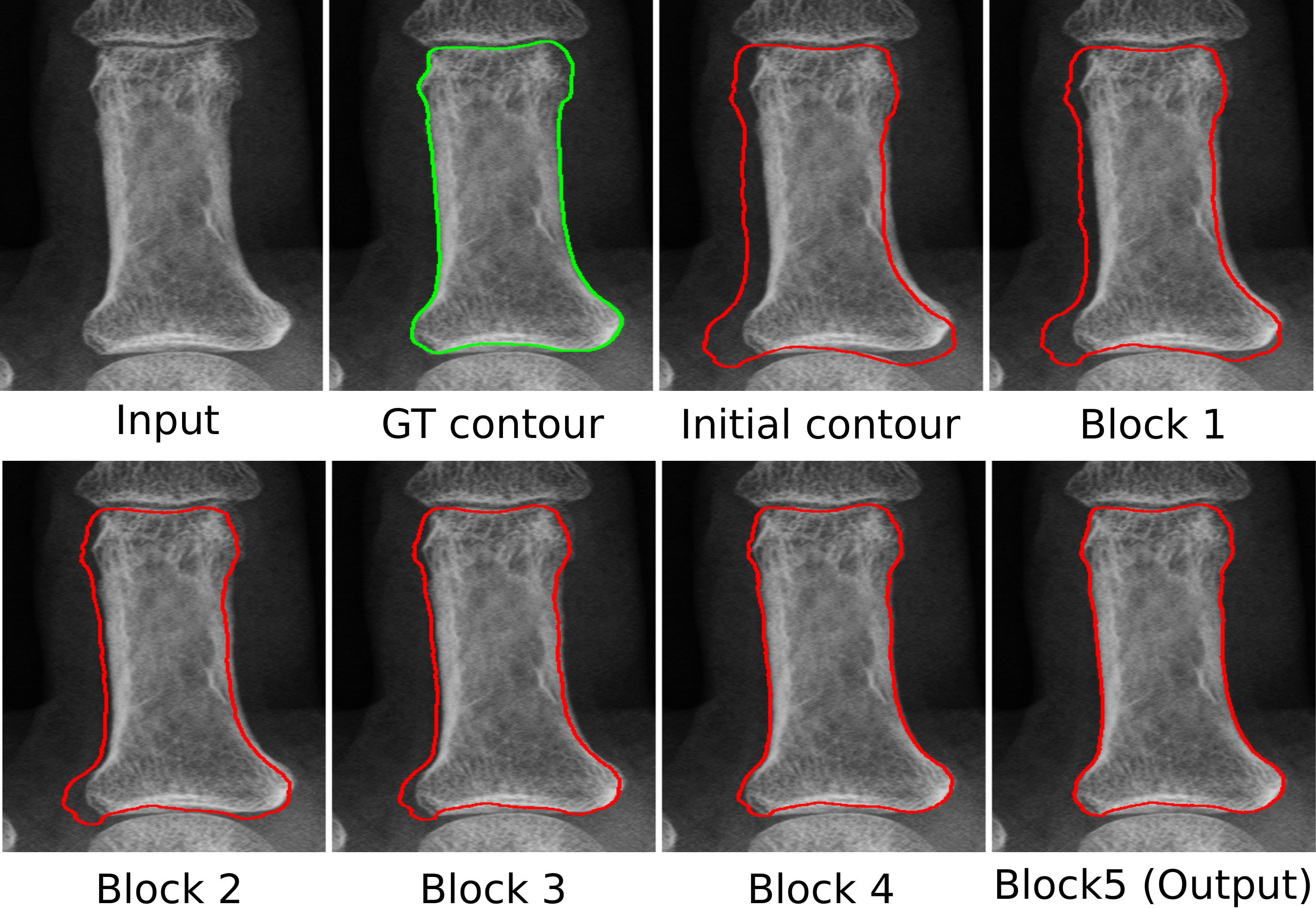}
    \end{center}
    \caption{
    \textbf{Visualization of the contour evolution process}. The red lines are the contours after each GCN block in CTN. It shows how CTN gradually moves the initial contour to the correct location.}
    \label{fig:contour_evolve}
\end{figure}

The results in Table~\ref{table:compare} show that our method significantly outperforms both MorphACWE and MorphGAC. Specifically, on average we achieve 15.63\% higher IoU and 20.94 pixels less HD than MorphGAC, the better of the two. The visualizations of segmentation results in Fig.~\ref{fig:seg_compare} confirm that these two approaches cannot provide satisfactory segmentation accuracy, especially when the boundary of such structures is not clear, e.g., lung segmentation. We posit that the inferior performance of ACM-based methods is owing to two factors: 1) the gradient-based energy function is not suitable for objects without clear boundary, 2) optimizing the energy function on single image often encounters local minima (i.e., causing segmentation leakage). In contrast, CTN optimizes shape and appearance-based loss functions on an aggregated of the unlabeled dataset to achieve high robustness.  Fig.~\ref{fig:contour_evolve} shows the evolution process of the CTN contour on a phalanx image.

\subsubsection{Comparison with one-shot methods}
We also compare with two representative one-shot segmentation methods: CANet~\cite{zhang2019canet}~\footnote{\url{https://github.com/icoz69/CaNet}} and Brainstorm~\cite{zhao2019data}~\footnote{\url{https://github.com/xamyzhao/brainstorm}}. CANet is trained on the PASCAL VOC 2012 dataset and can segment unseen objects by referring to the support set (the exemplar). Brainstorm tackles the one-shot segmentation problem by learning both spatial and appearance transformations between images in a dataset and further synthesizes image-label pairs to train the segmentation model. We follow their procedures to process images in our datasets. For all one-shot methods, including ours, we use the same exemplar as the one-shot data.

\setlength{\tabcolsep}{1.2mm}
\begin{figure}
    \begin{center}
    	\includegraphics[width=\linewidth]{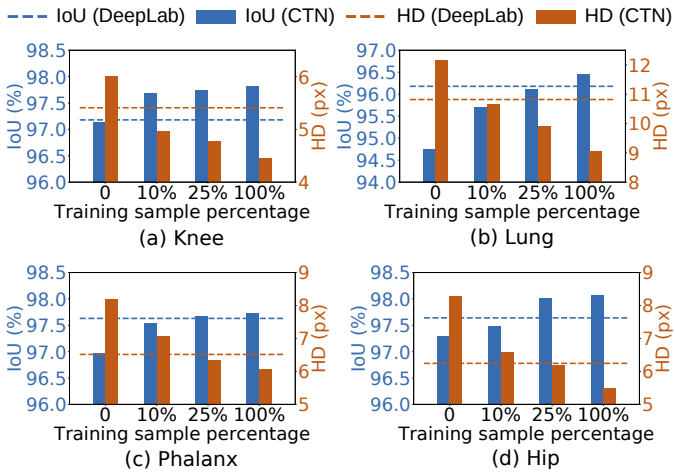}
    \end{center}
    \caption{\textbf{Using different number of human corrections to finetune the one-shot model}. We test the performance of the human-in-the-loop mechanism with 0, 10\%, 25\% and 100\% corrected training samples, respectively (``0'' means no finetuning). Our performance with 25\% training samples generally outperforms DeepLab using 100\% samples.}
    \label{fig:more_label}
\end{figure} 

As shown in Table~\ref{table:compare}, CANet achieves only 48.98\% IoU on average. We speculate that the poor performance is caused by the domain gap between natural images and medical images. Brainstorm achieves better performances with an average IoU and HD of 82.46\% and 36.71, respectively. This is still significantly lower than CTN, of which the average IoU and HD are 96.58\% and 8.66, respectively.
Fig.~\ref{fig:seg_compare} shows that while Brainstorm is able to segment the object's overall structure, it has low accuracy on the segmentation boundaries. 

\subsubsection{Comparison with fully supervised methods}
We also evaluate the performance of three fully supervised methods on our datasets: 
UNet~\cite{ronneberger2015u}~\footnote{\url{https://github.com/milesial/Pytorch-UNet}}, DeepLab-v3+~\cite{chen2018encoder}~\footnote{\url{https://github.com/jfzhang95/pytorch-deeplab-xception}} and HRNet-W18~\cite{wang2020deep}~\footnote{\url{https://github.com/HRNet/HRNet-Semantic-Segmentation}}. 
We train each of them with all available training data, \ie, 100 knee images, 124 lung images, 100 phalanx images, and 100 hip images, respectively.
Post-processing procedures are excluded for fair comparison.

CTN trained with only one exemplar performs comparably with the fully supervised UNet, and slightly falls behind DeepLab, the best of the baseline methods, by 0.58\% in IoU and 1.41 pixel in HD, respectively.
These results suggest that with only one exemplar, CTN can compete head-to-head with very strong fully supervised baselines. We note that since these fully supervised methods predict segmentation labels at pixel-level, the topology of the segmentation is not guaranteed, \eg, small isolated lung masks in Fig.~\ref{fig:seg_compare}. In contrast, CTN is able to retain the topology. Moreover, we will demonstrate in Section~\ref{subsec:more_label} that with minimal human feedback, CTN can even outperform fully supervised models. 

\setlength{\tabcolsep}{1.4mm}
\begin{table}[t]
\caption{\textbf{Using more unlabeled images in training.} We expand the training set of knee and phalanx from 100 to 500 images to examine our method's ability in exploiting unlabeled data. Both cases use only one exemplar.}
\begin{center}
	\begin{tabular}{c|cc|cc|cc}
		\hline
		\multirow{2}{*}{\begin{tabular}[c]{@{}c@{}}Unlabeled\\ images\end{tabular}} & \multicolumn{2}{c|}{Knee} & \multicolumn{2}{c|}{Phalanx}  & \multicolumn{2}{c}{Hip} \\ \cline{2-7} 
		& IoU(\%)          & HD(px)         & IoU(\%)           & HD(px)   & IoU(\%)           & HD(px)              \\ \hline
		100         & 97.32      & 6.01        & 96.96       & 8.19       & 97.29 & 8.27\\ 
		500         & 97.53      & 5.73      & 97.33       & 6.96       & 97.37   & 7.97\\ \hline
	\end{tabular}
\end{center}
\label{table:more_unlabel}
\end{table}


\setlength{\tabcolsep}{1.4mm}
\begin{table*}[ht]
\caption{\textbf{Ablation study}. The three losses of CTN are removed individually to evaluated their impacts on the segmentation performance.}
\label{table:ablation}
\begin{center}
\begin{tabular}{ccc|cc|cc|cc|cc|cc}
\hline
\multirow{2}{*}{$L_{perc}$} & \multirow{2}{*}{$L_{bend}$} & \multirow{2}{*}{$L_{edge}$} & \multicolumn{2}{c|}{Knee} & \multicolumn{2}{c|}{Lung} & \multicolumn{2}{c|}{Phalanx} & \multicolumn{2}{c|}{Hip} & \multicolumn{2}{c}{\textbf{Mean}} \\ \cline{4-13} 
                            &                              &                             & IoU (\%)          & HD (px)         & IoU (\%)          & HD (px)         & IoU (\%)           & HD (px)          & IoU (\%)          & HD (px)         & IoU (\%)          & HD (px)         \\ \hline
                            & $\checkmark$                 & $\checkmark$                & 94.62	&	8.28	&	87.45	&	26.51	&	94.01	&	15.80	&	92.90	&	16.58	&	92.24	&	16.79\\ 
$\checkmark$                &                              & $\checkmark$                & \textbf{97.49}	&	\textbf{5.87}	&	84.93	&	36.74	&	94.24	&	26.13	&	94.53	&	13.91	&	92.80	&	20.66\\ 
$\checkmark$                & $\checkmark$                 &                             & 94.43	&	11.90	&	93.00	&	16.22	&	96.45	&	9.84	&	96.61	&	9.92	&	95.12	&	11.97 \\ \hline
$\checkmark$                & $\checkmark$                 & $\checkmark$                & 97.32	&	6.01	&	\textbf{94.74}	&	\textbf{12.17}	&	\textbf{96.96}	&	\textbf{8.19}	&	\textbf{97.29}	&	\textbf{8.27}	&	\textbf{96.58}	&	\textbf{8.66} \\ \hline
\end{tabular}
\end{center}
\end{table*}

\subsection{Incorporating human corrections}
\label{subsec:more_label}

In this section, we validate the effectiveness of the proposed human-in-the-loop mechanism by simulating manual corrections of wrong segmentation by an annotator. Specifically, we assume that the annotator tends to correct more severe errors with higher priority. To simulate this behavior, we first segment the unlabeled training images using the one-shot trained model and calculate their HD to the ground-truth segmentation (which is not used in training). Then, we select the worst $n\%$ images as candidates for correction. For each predicted contour in these images, we calculate its point-wise L2 distances to the ground-truth and mark vertices with distances larger than 3 pixels as errors. We group consecutive error vertices into segments and use the corresponding ground-truth vertices as corrections. Under this setting, we conduct human-in-the-loop training using corrections of 10\%, 25\% and 100\% training images, respectively.

Fig.~\ref{fig:more_label} shows the performances of the original one-shot model and three human-in-the-loop finetuned models. We observe that our model consistently improves with more corrections. Specifically, using 10\% corrections, the mean IoU is improved from 96.58\% to 97.10\% and the mean HD is reduced from 8.66 to 7.32, respectively. When using 25\% corrections, CTN can outperform DeepLab, (IoUs of 97.38\% vs. 97.16\%, and HDs of 6.81 vs. 7.25). With corrections on all training samples, CTN further reaches an IoU of 97.52\% and a HD of 6.27. We also stress that the effort of our human-in-the-loop correction of unlabeled training samples is significantly lower than annotating them from scratch (as required by fully supervised methods), as only partial corrections are needed. Thus, these results indicate that on all 4 evaluated tasks, CTN with the human-in-the-loop mechanism can achieve superior performance than fully supervised methods and require considerably less annotation effort.

Knowing that human-in-the-loop fine-tuning improves the overall segmentation performance, we further investigate if the fine-tuned CTN may produce degraded performance on individual cases compared to the one-shot CTN. On the hip dataset, we found that on 178 out of 200 testing images, the IoU improved after fine-tuning using 25\% corrections (average IoU from 97.27\% to 98.12\%). On the other 22 testing images, the IoU degraded (average IoU from 97.48\% to 97.06\%). Overall, the average IoU of all 200 images increased from 97.3\% to 98.0\%. The results show that in the majority of the cases (89\%), fine-tuned CTN improves the segmentation performance by a noticeable IoU gap (0.85\%). While in some cases (11\%), the performance degrades, the degradation is on average smaller (IoU gap 0.42\%) than the improvement. 

\subsection{Training with more unlabeled data}
\label{subsec:more_data}
Another advantage of CTN is that it can utilize more unlabeled data (which are often easy to obtain) in training to improve its performance. To evaluate the impact of more unlabeled data by expanding the unlabeled training sets of knee, hip and phalanx from 100 images to 500 images, with the exemplar unchanged. We do not conduct this experiment on the lung dataset, because there is no additional images available in the JSRT dataset.

As shown in Table~\ref{table:more_unlabel}, by increasing the number of unlabeled images from 100 to 500, the performance improves on average by 0.22\% in IoU and 0.6 in HD. Among the three datasets, the improvement on the phalanx dataset is the largest. Phalanx dataset has larger appearance and shape variations than hip and knee, since it contains bones from 5 fingers. We hypothesize that CTN needs more training samples to fully capture the large appearance and shape variations.

\subsection{Ablation study on the proposed losses}
\label{sec:loss_ablation}
We conduct an ablation experiment to evaluate the effectiveness of the three employed losses, namely the contour perceptual loss $L_{perc}$, the contour bending loss $L_{bend}$, and the edge loss $L_{edge}$. The results are summarized in Table~\ref{table:ablation}. The performance of CTN degrades if any loss is removed, with an average IoU decrease of 4.34\%, 3.78\%, and 1.46\% for $L_{perc}$, $L_{bend}$, and $L_{edge}$, respectively. This demonstrates the contributions of all three losses. An exception is the knee dataset when $L_{bend}$ is removed. Knee X-ray images share similar appearance features along the contour so that they can be segmented robustly with just the contour perceptual loss and edge loss.
Thus, adding contour bending loss leads to slightly lower performance in this particular scenario, where the IoU decreases from 97.49\% to 97.32\% and the HD increases from 5.87 to 6.01. 
We note that the changes (0.17\% for IoU and 0.14 for HD) are below the standard deviations of CTN on knee (0.67\% for IoU and 1.42 for HD).
Despite the exception on the knee dataset, such a regularization effect by the contour bending loss is generally desired to alleviate the worst-case scenarios and is proved useful in the other three datasets.

\begin{figure}[tbp]
	\begin{center}
		\includegraphics[width=\linewidth]{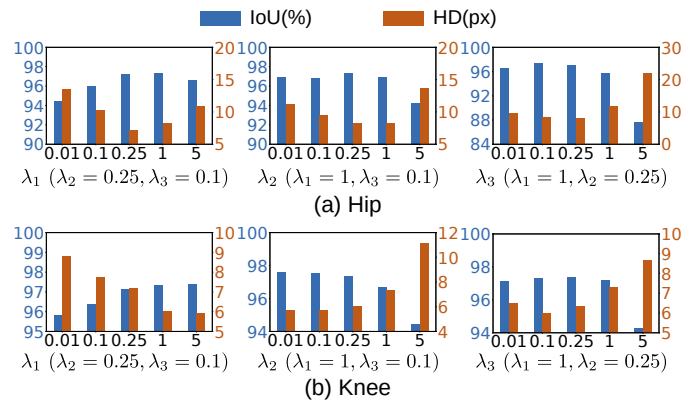}
	\end{center}
	\caption{\textbf{Using different loss weights to train CTN on the hip and the knee datasets.} Based on the original setting $\lambda_1=1$, $\lambda_2=0.25$, and $\lambda_3=0.1$, we change one of them each time and fix the other two.}
	\label{fig:loss_weight}
\end{figure}

To further understand the impact of the losses, we analyze CTN's sensitivity to the three loss weights, $\lambda_1$, $\lambda_2$ and $\lambda_3$.
Specifically, on both the knee and hip datasets, three experiments are conducted to evaluate the impact of varying the three loss weights individually while fixing the other two.
The CTN is trained and tested using 5 different values [0.01, 0.1, 0.25, 1, 5] for $\lambda_1$, $\lambda_2$ and $\lambda_3$.
The IoUs and HDs obtained using varying loss weights are reported in Fig.~\ref{fig:loss_weight}.
On the hip dataset, very small and large loss weights in general lead to degraded performance. 
On the knee dataset, larger $\lambda_1$ and smaller $\lambda_2$ achieve better performances. We posit that due to the distinct appearance, knees can be reliably segmented using the visual patterns (measured by $L_{perc}$) only, and strong shape regularization (measured by $L_{bend}$) degrades the performance by imposing unnecessary shape constraints.

We also compare the performances of CTN using L1 and L2 distances in the contour perceptual loss (Section \ref{sec:percep_loss}) on the hip dataset. The results show that using L2 distance results in degraded performance compared to using L1 distance, reporting an IoU of 96.82\% (compared to 97.29\%) and a HD of 12.41 (compared to 8.27). We note that the degradation in HD is more obvious than IoU, hypothetically owing to the forgiving nature of L2 distance to small errors.

\subsection{Analysis on failure cases}
\begin{figure}[t]
    \begin{center}
    	\includegraphics[width=0.9\linewidth]{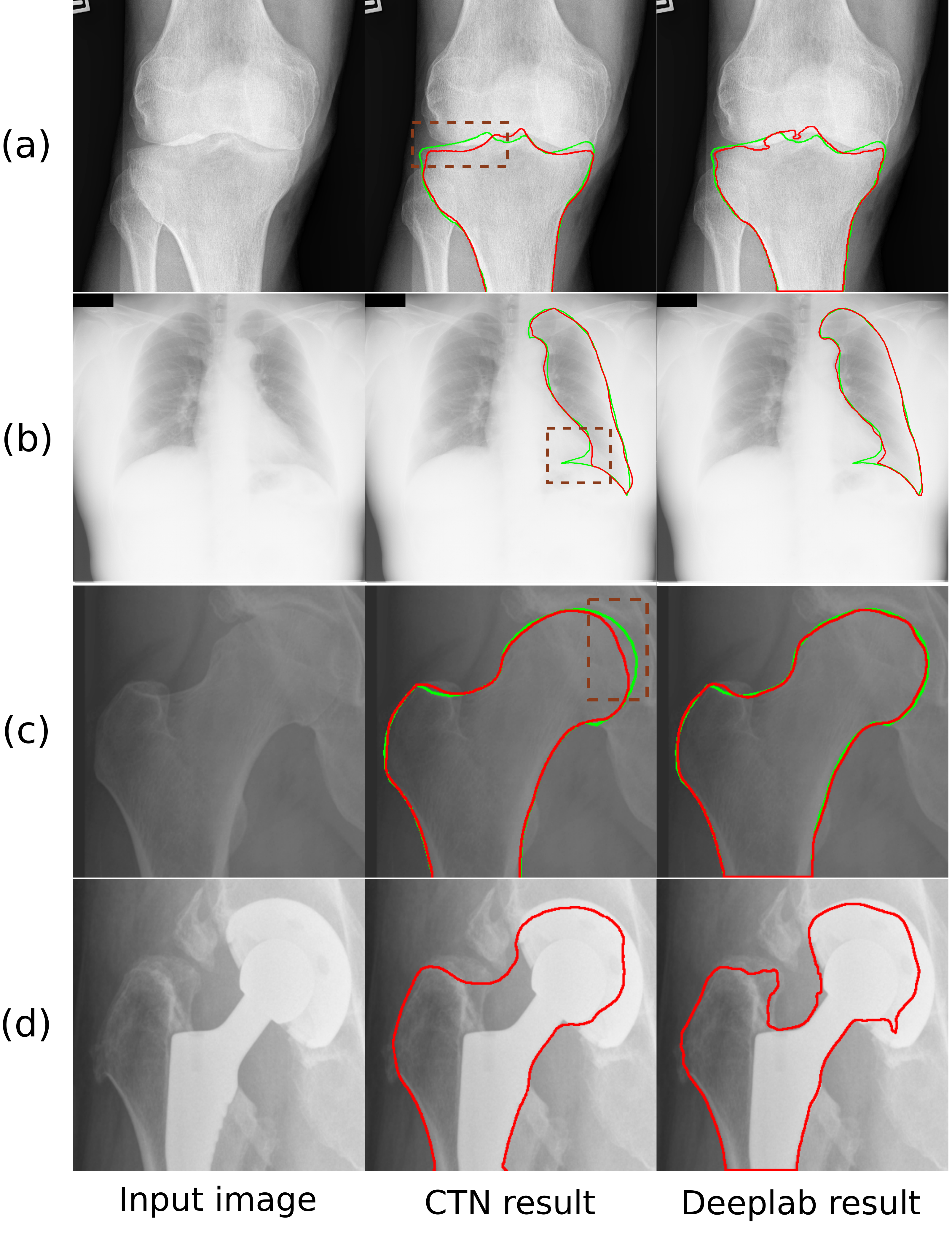}
    \end{center}
    \caption{\textbf{Typical failure cases}. (a) Intersected boundaries. (b) Acute shape change. (c) Blurry boundary. (d) Metal implant (this case is not in the test set). The green curves are the ground truth contours, the red curves are the predicted contours, and the dashed boxes show the wrong part of CTN predictions.}
    \label{fig:failure_cases}
\end{figure}

In Fig. \ref{fig:}, we show and examine a few typical failure cases to analyze the performance characteristics of CTN. 
Fig.~\ref{fig:failure_cases}(a) is a knee with severe osteoporosis, which significantly reduces the joint space and makes the tibia and femur bones overlap. CTN fails to segment the overlapped region properly. However, DeepLab also produces wrong segmentation on this challenging case. In Fig~\ref{fig:failure_cases}(b), the acute change of lung shape differs from the mean shape, and CTN mis-segments this part and produces a result closer to mean shape. Although DeepLab also mis-segments the same part, its result is closer to the ground truth than CTN. 
Fig.~\ref{fig:failure_cases}(c) is a hip with severe osteoporosis, similar to Fig.~\ref{fig:failure_cases}(a), where the joint space is reduced, making the bone boundary less recognizable. On this case, CTN produces wrong segmentation on the bone boundary affected by the osteoporosis, while DeepLab produces satisfactory results. 
Fig.~\ref{fig:failure_cases}(d) shows an extreme case of hip X-ray with total hip replacement (this case is not in our test since there is no ground truth segmentation). While there is no standard for correct segmentation on this particular case, we observe that the segmentation produced by CTN tends to follow the mean shape of a normal hip. In comparison, DeepLab tends to produce segmentation results following the edges in the image.

\subsection{Analysis on the behaviors of CTN}
\subsubsection{Robustness to detected ROIs}
As a prerequisite of CTN, ROI detection is an important step to help reduce the contour searching space. 
Therefore, the performance of CTN also depends on the accuracy of ROI detection.
To evaluate the influence of ROI detection, we conduct an experiment to compare the performances of CTN when using automatic ROIs and manually perturbed ROIs. 
Specifically, three offsets are imposed on the bounding boxes of hip ROIs to perturb their locations, $\Delta x$, $\Delta y$ and $\Delta \theta$, denoting the translation on x-axis and y-axis, and the rotation around the ROI center, respectively. 
We randomly generate $-5px\leq \Delta x, \Delta y \leq 5px$ and $-5^{\circ}\leq \Delta \theta \leq 5^{\circ}$ to simulate ROIs produced with certain landmark detection errors. Note that in our experiment, this perturbation is added on the automatically detected ROIs, which already contains errors from the ROI detector. 
We test the model trained without ROI perturbation on perturbed ROIs, to examine CTN's robustness to ROI localization errors unseen in the training data.
Table~\ref{table:perturb_roi} summarizes the testing results. 
With all three perturbations, the IoU dropped by 0.78\% and the HD increased by 1.88 px, indicating that the performance of CTN can be affected by ROI localization errors. However, the performance degradation is relatively small, \ie{}, comparable to the standard deviations (IoU 0.72\% and HD 3.06 px), indicating that CTN holds a good robustness against the perturbations. 
We evaluated DeepLab under the same ROI perturbation settings, and observed similar performance degradation, \ie{} IoU by 0.71\% and HD by 1.2 px.

\setlength{\tabcolsep}{2mm}
\begin{table}[t]
\caption{\textbf{Testing CTN on perturbed hip ROIs.} We manually modify the location of ROIs with offsets on x-axis, y-axis, and rotation.}
\label{table:perturb_roi}
\begin{center}
\begin{tabular}{ccc|cc}
\hline
$\Delta x$(px) & $\Delta y$(px) & $\Delta \theta$($^{\circ}$) & IoU (\%) & HD (px) \\ \hline
0              & 0              & 0                           & 97.29   & 8.27   \\
5              & 0              & 0                           & 96.90   & 8.76   \\
0              & 5              & 0                           & 96.91   & 8.88   \\
0              & 0              & 5                           & 96.62   & 9.79   \\
5              & 5              & 5                           & 96.51   & 10.15  \\ \hline
\end{tabular}
\end{center}
\end{table}

\begin{figure*}[t]
\begin{center}
\includegraphics[width=0.75\linewidth]{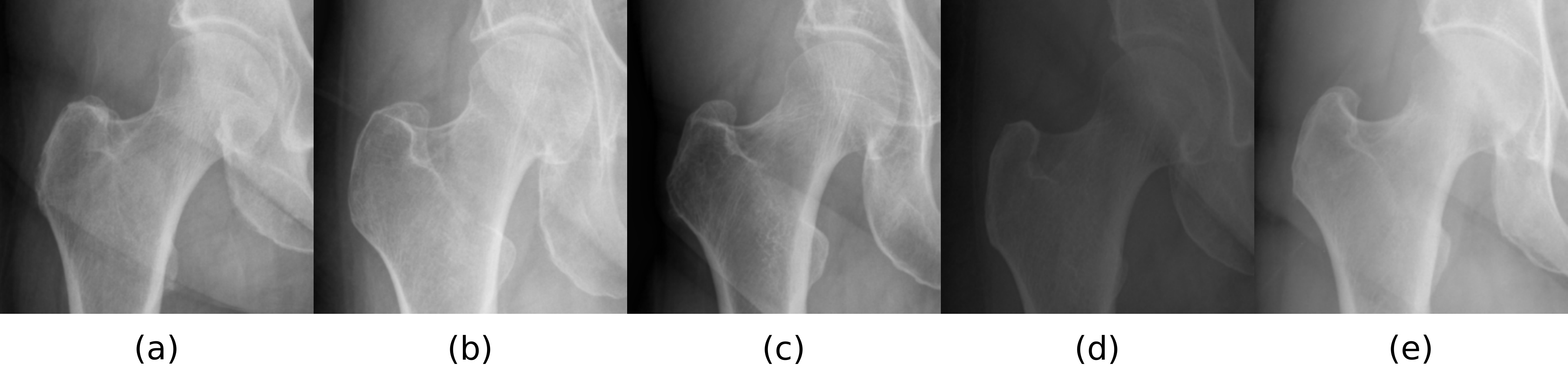}
\end{center}
\caption{\textbf{Five exemplar images from the hip training set.} (a) The exemplar image automatically selected using the proposed method. (b)-(e) Four randomly selected exemplars.}
\label{fig:exemplars}
\end{figure*}

\subsubsection{Impact of the selection of the exemplar image}
Since the exemplar image is the main source of supervision signal, the selection of the exemplar image may be critical to the generalizability of CTN. In Section~\ref{sec:dataset}, we propose to select the image with the minimum average VGG distance to other training images as the exemplar. In this section, we conduct experiments on the hip dataset to evaluate CTN's performance using four randomly selected exemplars and compare them with the automatically selected exemplar. The five exemplar images are shown in Fig.~\ref{fig:exemplars}, and their resulting performances are reported in Table~\ref{table:exemplar}. We can observe that the automatically selected exemplar Fig.~\ref{fig:exemplars}(a) based on VGG distance results in better performance than the randomly selected ones. We also observe that the performance of CTN is not always correlated with the VGG distance. For example, the exemplar Fig.~\ref{fig:exemplars}(d) with larger distance produces a better CTN model than Fig.~\ref{fig:exemplars}(e) with smaller distance. We note that even with randomly selected exemplars, CTN consistently outperforms previous one-shot segmentation methods, \eg{} Brainstorm.

\setlength{\tabcolsep}{2mm}
\begin{table}[t]
\caption{Using five different exemplars to train CTN on the hip dataset. The average VGG distance is calculated by averaging the L2 distances from the exemplar image to all other training images in the VGG feature space.}
\label{table:exemplar}
\begin{center}
\begin{tabular}{c|ccc}
\hline
Exemplar image & Avg. VGG distance & IoU(\%) & HD(px) \\ \hline
\textbf{Fig.~\ref{fig:exemplars}(a)}     & $2.98\times10^5$       & \textbf{97.29}    & \textbf{8.27}    \\
Fig.~\ref{fig:exemplars}(b)              & $3.51\times10^5$       & 96.88    & 9.39    \\
Fig.~\ref{fig:exemplars}(c)              & $3.32\times10^5$       & 96.08    & 11.92   \\
Fig.~\ref{fig:exemplars}(d)              & $4.83\times10^5$       & 96.77    & 9.36    \\
Fig.~\ref{fig:exemplars}(e)              & $3.20\times10^5$       & 94.57    & 15.37   \\ \hline
\end{tabular}
\end{center}
\end{table}

\begin{table}[t]
\caption{Using different numbers of GCN blocks to train CTN on the hip dataset.}
\label{table:gcn_blocks}
\begin{center}
\begin{tabular}{c|ccccc}
\hline
Num. of GCN blocks & 1     & 3     & 5     & 7     & 9     \\ \hline
IoU (\%)          & 97.10 & 97.11 & 97.29 & 96.91 & 96.67 \\
HD (px)           & 8.51  & 8.44  & 8.27  & 8.88  & 9.29  \\ \hline
\end{tabular}
\end{center}
\end{table}

\subsubsection{Impact of the number of GCN block iterations}
In this section, we evaluate the impact of the number of GCN block iterations by training and testing the CTN with 1, 3, 5, 7 and 9 GCN block iterations on the hip dataset. The results of this analysis are summarized in Table~\ref{table:gcn_blocks}. It shows that as the number of GCN blocks increases from 1 to 5, the performance improves from IoU 97.10\% to 97.29\% and HD 8.51 px to 8.27 px, respectively. It demonstrates that by stacking multiple GCN blocks, the later GCN block can further correct the segmentation errors produced by the earlier GCN blocks, which is beneficial to the final performance. However, the performance starts to slightly degrade when the number of GCN blocks increases over 5. We posit that the increased number of layers in the CTN caused by the additional GCN blocks make the network more difficult to train, which contributes to the performance degradation. 

\subsubsection{Computational efficiency}
We analyze the computational efficiency of CTN and compare it with other learning-based segmentation methods. Table~\ref{table:model_eff} summarizes the number of parameters, the number of float-point operations (FLOPs) and frames per second (FPS). All evaluations are conducted on the hip dataset with a Nvidia GTX 1080Ti GPU. While the computational efficiency varies significantly among the evaluated methods (\eg{} number of parameters from 1.78M to 59.34M, FLOPs from 4.67G to 32.99G, FPS from 9.86 to 28.57), all methods report sufficient speed (above 9 FPS) for off-line image analysis tasks. A few methods, including CTN, measure above 15 FPS, which is the common fluoroscopic imaging frame rate, showing potential applicability on real-time image analysis tasks.

\setlength{\tabcolsep}{4mm}
\begin{table}
\caption{\textbf{Model efficiency of learning-based methods.} We compare the number of parameters, the number of float-point operations (FLOPs), and the inference FPS of all learning-based methods.}
\begin{center}
\begin{tabular}{c|ccc} \hline
Methods & \# of Params & FLOPs & FPS \\ \hline
CANet           &    19.01M                        &    27.42G             &   20.77           \\
Brainstorm      &      1.78M                      &    7.55G             &    15.44          \\
UNet            & 13.40M                       & 30.65G           & 28.57        \\
DeepLab         & 59.34M                      & 22.55G           & 18.18        \\
HRNet           & 9.64M                       & 4.67G            & 9.86        \\ 
CTN             & 42.26M                      & 32.99G           & 15.39       \\\hline
\end{tabular}
\end{center}
\label{table:model_eff}
\end{table}

\section{Discussion and conclusion}
In this paper, we presented CTN, a one-shot segmentation method that can be trained using one labeled exemplar and a set of unlabeled images.
We demonstrated that by properly exploiting the regularized nature of anatomical structures, CTN trained with one labeled data (exemplar) can compete head-to-head with fully supervised methods trained with abundant labeled data.
A key assumption of our work is that the same anatomy have similar shape and visual patterns in different images.
Based on this assumption, CTN employs a semi-supervised training strategy with losses that measures the similarity between the segmentation from unlabeled images and the exemplar. 
A key difference between CTN and most existing segmentation methods (one-shot and supervised) is that CTN models segmentation as contour and learns the contour evolution behavior.
Using contour representation makes it possible to directly compare the shapes of segmentation results, as well as measure the similarity of visual appearance along the segmentation boundary.
We have shown that shape similarities can be measured using TPS bending energy of the two contours and used as training loss, which is sensitive to acute shape changes and is suitable for imposing shape regularization to prevent irregular segmentation.
Visual pattern similarities of two contours can be evaluated by comparing the features of corresponding vertices in the ImageNet trained VGG feature space. Since the VGG is trained on ImageNet, its feature is salient to the structure and insensitive to low level image variations, which is ideal for comparing the visual similarity of two segmentation contours.

Section \ref{subsec:more_data} and \ref{subsec:more_label} demonstrate that the performance of CTN can be further improved in two ways, training with more unlabeled data and incorporating human-in-the-loop corrections, respectively. By using more unlabeled training data, without addition annotation effort, CTN can reach the performance of the state-of-the-art supervised segmentation methods (e.g., DeepLab). The human-in-the-loop correction is high labor cost-effective, i.e., the annotator only needs to draw the mis-segmented partial contour. As shown in Fig.\ref{fig:more_label}, with human-in-the-loop, CTN can outperform supervised methods by a large margin, especially on HD. For one-shot learning methods to be useful in clinical applications, especially the accuracy demanding ones, the capability to effectively incorporate human-in-the-loop corrections to boost performance is a critical feature. However, most existing one-shot methods fail to provide such mechanism.

We recognize that CTN also has its limitations. The success of CTN is achieved by heavily exploiting the assumption that the target anatomical structure has similar shape and appearance in different images. If the anatomical structure has significant difference from the exemplar in shape and/or appearance (e.g., caused by pathology), the contour bending loss and contour perceptual loss may provide misinformed guidance to CTN and we expect the performance of CTN to degrade. This limitation can be partially addressed by the human-in-the-loop mechanism with certain manual correction efforts. Another limitation of CTN is that it can only utilize one exemplar and does not support few-shot learning scenarios. This is mainly because the contour bending loss and contour perceptual loss are calculated pair-wise between the exemplar and the unlabeled images. Future research could investigate the extension of CTN to few-shot learning scenario via group-wise loss calculation. In addition, the extension of CTN to 3D segmentation might prove an important area for future research. Unlike FCN-based segmentation methods, which can be directly applied on 3D tasks by using 3D convolutions, extending the 2D contour-based formulation of CTN to a 3D surface-based formulation requires is non-trivial and warrants further investigation.



{\small
\bibliographystyle{ieee_fullname}
\bibliography{ctn.bib}
}

\end{document}